\documentclass[sigconf]{acmart}
\AtBeginDocument{%
  }

\usepackage{multirow}
\usepackage{hyperref}

\copyrightyear{2025}
\acmYear{2025}
\setcopyright{acmlicensed}\acmConference[MM '25]{Proceedings of the 33rd
ACM International Conference on Multimedia}{October 27--31, 2025}{Dublin,
Ireland}
\acmBooktitle{Proceedings of the 33rd ACM International Conference on
Multimedia (MM '25), October 27--31, 2025, Dublin, Ireland}
\acmDOI{10.1145/3746027.3755489}
\acmISBN{979-8-4007-2035-2/2025/10}

\begin{document}

\title{HCCM: Hierarchical Cross-Granularity Contrastive and
Matching Learning for Natural Language-Guided Drones}

\author{Hao Ruan}
\affiliation{%
  \department{Department of Artificial Intelligence,}
  \institution{Xiamen University}
  \city{Xiamen}
  \country{China}
}
\email{ruanhao@stu.xmu.edu.cn}

\author{Jinliang Lin}
\affiliation{%
  \department{Department of Artificial Intelligence,}
  \institution{Xiamen University}
  \city{Xiamen}
  \country{China}
}
\email{jinlianglin@stu.xmu.edu.cn}

\author{Yingxin Lai}
\affiliation{%
  \department{Department of Artificial Intelligence,}
  \institution{Xiamen University}
  \city{Xiamen}
  \country{China}
}
\email{laiyingxin2@gmail.com}

\author{Zhiming Luo}
\authornote{Corresponding author}
\affiliation{%
  \department{Key Laboratory of Multimedia Trusted Perception and Efficient Computing, Ministry of Education of China,}
  \institution{Xiamen University}
  \city{Xiamen}
  \country{China}
}
\email{zhiming.luo@xmu.edu.cn}

\author{Shaozi Li}
\affiliation{%
  \department{Fujian Key Laboratory of Big Data Application and Intellectualization for Tea Industry,}
  \institution{Wuyi University}
  \city{Wuyishan}
  \country{China}
}
\email{szlig@xmu.edu.cn}

\renewcommand{\shortauthors}{Hao Ruan, Jinliang Lin, Yingxin Lai, Zhiming Luo, and Shaozi Li}

\begin{abstract}
Natural Language-Guided Drones (NLGD) offer a novel and flexible interaction paradigm for tasks such as target matching and navigation. However, the wide field of view and complex compositional semantic relationships inherent in drone scenarios place greater demands on visual language understanding. First, mainstream Vision-Language Models (VLMs) primarily focus on global feature alignment and lack fine-grained semantic understanding. Second, existing hierarchical semantic modeling methods rely on precise entity partitioning and strict containment relationship constraints, which limits their effectiveness in complex drone environments. To address these challenges, we propose the Hierarchical Cross-Granularity Contrastive and Matching learning (HCCM) framework, comprising two core components:
1) Region-Global Image-Text Contrastive Learning (RG-ITC). Avoiding precise scene entity partitioning, RG-ITC models hierarchical local-to-global cross-modal semantics by contrasting local visual regions with global text semantics, and vice versa. 2) Region-Global Image-Text Matching Learning (RG-ITM). Instead of relying on strict relationship constraints, this component evaluates local semantic consistency within global cross-modal representations, improving the comprehension of complex compositional semantics.
Furthermore, drone scenario textual descriptions are often incomplete or ambiguous, destabilizing global semantic alignment. To mitigate this, HCCM incorporates a Momentum Contrast and Momentum Distillation (MCD) mechanism, enhancing alignment robustness.
Extensive experiments on the GeoText-1652 benchmark demonstrate HCCM significantly outperforms existing methods, achieving state-of-the-art Recall@1 scores of 28.8\% (image retrieval) and 14.7\% (text retrieval). Moreover, HCCM exhibits strong zero-shot generalization on the unseen ERA dataset, achieving 39.93\% mean recall (mR), surpassing evaluated fine-tuned models. These results highlight the effectiveness and robustness of HCCM across diverse scenarios. Our implementation is available at \emph{\url{https://github.com/rhao-hur/HCCM}}.
\end{abstract}

\ccsdesc[500]{Information systems~Image search}
\ccsdesc[300]{Computing methodologies~Visual content-based indexing and retrieval}
\keywords{Natural Language-Guided Drones, Cross-Modal Retrieval, Compositional Semantics}


\maketitle

\begin{figure}[h]
  \centering
  \includegraphics[width=0.95\linewidth]{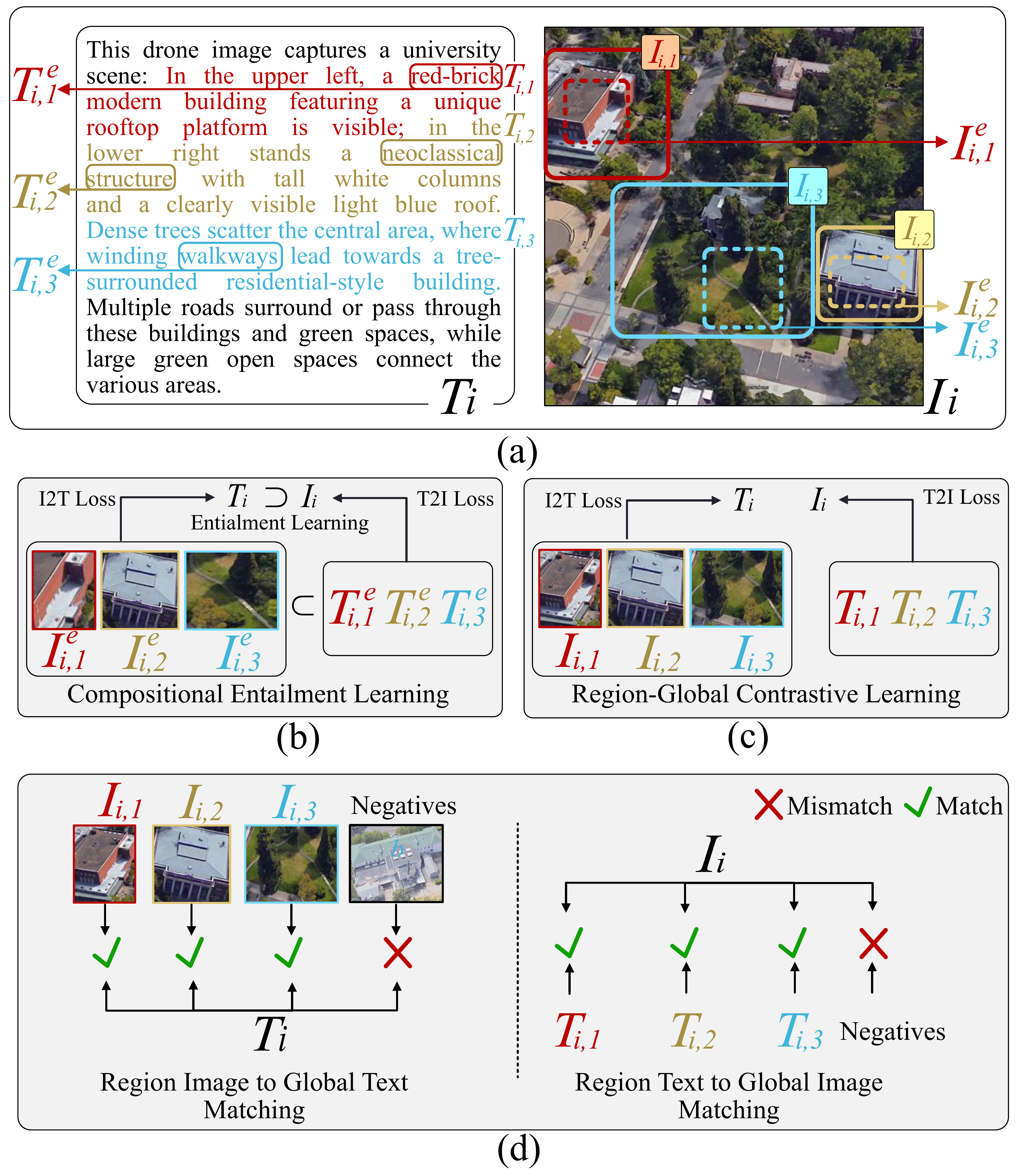}
\caption{Comparison of hierarchical vision-language modeling approaches.
    \textbf{(a)} Example input pair ($I_i, T_i$) demonstrating multiple semantic levels. The global ($I_i, T_i$), local region ($I_{i,k}$, solid box; $T_{i,k}$, sentence), and entity ($I_{i,k}^e$, dashed box; $T_{i,k}^e$, phrase).
    \textbf{(b)} Existing approach \cite{pal2024compositional} models entity-level hierarchies ($I_{i,k}^e, T_{i,k}^e$) using strict part-whole entailment.
    \textbf{(c)} Proposed RG-ITC contrasts local region features ($I_{i,k}$ or $T_{i,k}$) against the complementary global representation ($T_i$ or $I_i$) for cross-granularity semantic alignment.
    \textbf{(d)} Proposed RG-ITM assesses region-global cross-modal consistency by matching local features ($I_{i,k}$ or $T_{i,k}$) with the complementary global representation ($T_i$ or $I_i$) (Match \checkmark{} / Mismatch $\times$).
}
  \label{fig:scene_explanation}
\end{figure}

\section{Introduction}

In recent years, the application of Unmanned Aerial Vehicles (UAVs) has expanded from basic image acquisition to include complex tasks such as agricultural monitoring \cite{gago2015uavs, kim2019sustainable, tripicchio2015towards, wardihani2018real}, target tracking \cite{zhang2022visible, chen2022uav, nousi2019embedded}, and cross-view target matching \cite{zheng2020university, lin2022joint}. Among these, cross-view target matching has emerged as a crucial task, aiming to locate targets by matching images captured from different perspectives (e.g., UAV, satellite, ground), which is often formulated as an image retrieval problem. However, relying solely on visual queries encounters challenges in cross-view target matching: performance is susceptible to variations in illumination, weather, and viewpoint changes, leading to degradation \cite{wang2024Muse, lin2024self}. Furthermore, visual queries may not always be available in practical applications. Consequently, leveraging Natural Language-Guided Drones (NLGD) for target matching has emerged as a significant research direction, owing to their flexible querying approach and integrated language understanding capabilities \cite{chu2024geotext1652}.


Recent research in Natural Language-Guided Drones (NLGD) shows significant progress. To support research in the NLGD task, Chu et al. \cite{chu2024geotext1652} introduced a large-scale NLGD dataset, GeoText-1652, and defined two core subtasks: \textbf{UAV Text Navigation} (text-guided UAV positioning) and \textbf{UAV View Target Localization} (matching descriptions to UAV views for target identification). They employed Vision-Language Models (VLMs) \cite{radford2021clip, chen2020uniter, li2022blip, li2021albef, zeng2021xvlm, zhang2025poison} with contrastive learning to align global image-text representations in a shared embedding space. Concurrently, Huang et al. \cite{Huang2024vcsr} introduced the ERA and UDV datasets for NLGD, but explored non-VLM approaches. They utilized Convolutional Neural Networks (CNNs) \cite{wang2017aggregating, radoi2019multilabel, zhang2019vaa} and Bidirectional Gated Recurrent Units (Bi-GRUs) \cite{cho2014learning} for visual-language encoding and developed the Text-Guided Visual Information Reasoning (TGVIR) mechanism for fine-grained cross-modal semantic alignment.

However, NLGD task requires handling queries with compositional semantics, while current methods \cite{chu2024geotext1652,Huang2024vcsr} often exhibit \textbf{poor generalization for compositional understanding} and \textbf{fail to grasp cross-granularity semantic hierarchies}. As illustrated in Figure \ref{fig:scene_explanation}(a), let $I_i, T_i$ be the global image/text, $I_{i,k}, T_{i,k}$ the local region image/text (solid box/sentence), and $I_{i,k}^e, T_{i,k}^e$ the fine-grained entity image/text (dashed box/phrase). A global image $I_i$ typically contains multiple entity-level semantic regions (e.g., $I_{i,1}^e, I_{i,2}^e, I_{i,3}^e$) corresponding to entity descriptions (e.g., $T_{i,1}^e$, $T_{i,2}^e$, $T_{i,3}^e$) within the global text $T_i$. The compositional interplay of these regions defines the scene's semantics. Accurate scene distinction or instruction execution requires understanding the compositional roles of local regions within the global context. VLMs relying on global semantic alignment, often lack this fine-grained understanding. Furthermore, traditional sequence models \cite{cho2014learning} generalize poorly when interpreting complex compositional relationships, particularly in longer texts \cite{liu2020compositional}.

Recognizing the need to model relationships across different granularities, some VLM-based methods explore specific cross-modal interactions. For instance, Pal et al. \cite{pal2024compositional} proposed Compositional Entailment Learning (Fig.~\ref{fig:scene_explanation}(b)), modeling part-whole hierarchies via cross-modal contrastive learning ($I_{i,k}^e \to T_i$, $T_{i,k}^e \to I_i$) within hyperbolic space. This approach leverages semantic entailment learning \cite{le2019inferring}, assuming more abstract entities ($I_{i,k}^e, T_{i,k}^e$) entail the global concrete concepts ($I_i, T_i$). Typically, textual descriptions tend to express more abstract concepts than images. Formally, A entails B is defined as $B \subset A$, implying intra-modal relations $ T_i \subset T_{i,k}^e$ and $I_i \subset I_{i,k}^e$, and inter-modal relations $I_{i,k}^e \subset T_{i,k}^e$ and $I_i \subset T_i$.
However, applying such strict, entailment-based hierarchies proves challenging for UAV bird's-eye views. UAV imagery frequently features complexly intertwined elements (e.g., the Z-shaped road in Fig.~\ref{fig:scene_explanation}(a)) and widely distributed similar elements (e.g., trees), resisting clear delineation into discrete entities suitable for rigid decomposition. Moreover, UAV scene descriptions often prioritize element co-occurrence and composition over strict semantic entailment. Consequently, the geometric constraints imposed by entailment learning (e.g., entailment cone loss \cite{ganea2018entailment}) may be overly restrictive for flexibly capturing the compositional semantics inherent in UAV scenario.

\begin{figure}[h]
  \centering
  \includegraphics[width=1\linewidth]{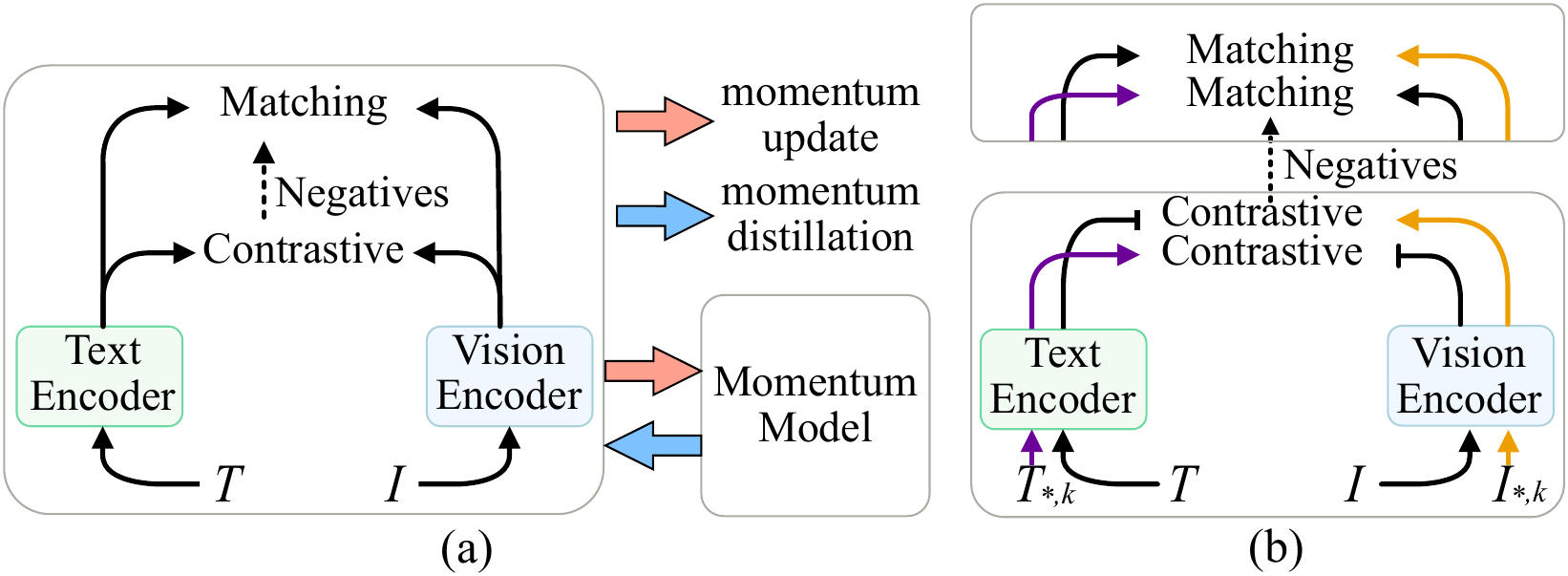}
\caption{Comparison of contrastive and matching learning frameworks. (a) Momentum-enhanced framework with update and distillation mechanisms for improved stability, built upon the standard bidirectional global alignment (left). (b) Proposed region-global framework combining global ($T, I$) and regional ($T_{*,k}, I_{*,k}$) information with unidirectional region-to-global contrastive learning.}
\label{fig:overall_compare}
\end{figure}

To address the above issues, we propose the \textbf{H}ierarchical \textbf{C}ross-Granularity \textbf{C}ontrastive and \textbf{M}atching Learning (HCCM) method. Building upon the standard cross-modal contrastive and matching learning framework \cite{zeng2021xvlm}, HCCM introduces Region-Global Image Text Contrastive Learning (RG-ITC) (Figure~\ref{fig:scene_explanation}(c)). Unlike methods relying on entity partitioning, RG-ITC models semantic associations across granularities, specifically linking unimodal local information (image region $I_{i,k}$ or text fragment $T_{i,k}$) with the corresponding global representation of the other modality (text $T_i$ and image $I_i$). This aims to capture the local-to-global cross-modal semantic hierarchical relationships within UAV scenarios. Furthermore, distinct from approaches modeling strict parent-child or part-whole relationships, we introduce Region-Global Image Text Matching Learning (RG-ITM) (Figure~\ref{fig:scene_explanation}(d)) to enhance the model's ability to discern the semantic consistency between local details and the global context across modalities. Specifically, it assesses whether the semantic content derived from a unimodal local region ($I_{i,k}$ or $T_{i,k}$) is consistent with the corresponding global representation of the other modality ($T_i$ or $I_i$). This process improves the model's comprehension and discrimination of complex spatial layouts and intertwined semantics.

However, directly applying this strategy in drone scenarios encounters certain limitations. Large-scale views often yield incomplete or ambiguous $T_i$, causing local-global alignment to amplify noise bias \cite{arpit2017closer}, impairing global performance.
To mitigate this, we introduce Momentum Contrast and Momentum Distillation (MCD), employing negative queues and soft targets respectively to stabilize global alignment and enhance interference resistance.
By combining RG-ITC, RG-ITM and MCD, our proposed HCCM can effectively improve the performance of VLM in UAV scenarios.

In summary, the main contributions of this paper are as follows:
\begin{enumerate} 
    \item A Hierarchical Cross-Granularity Contrastive and Matching Learning (HCCM) framework is presented to address insufficient fine-grained feature alignment and difficulty in modeling hierarchical relationships in Natural Language-Guided Drone tasks.
    
    \item Region-Global Image Text Contrastive Learning (RG-ITC) is designed to model cross-granularity hierarchies, and Region-Global Image Text Matching Learning (RG-ITM) is proposed to enhance composite semantic understanding.
    
    \item A Momentum Contrast and Momentum Distillation (MCD) strategy is introduced to mitigate noise amplification from incomplete text descriptions.
    
    \item Experiments on GeoText-1652 and ERA datasets validate the effectiveness and robustness of the proposed method.
\end{enumerate}

\section{Related Work}
\subsection{Vision and Language Navigation}
Using natural language descriptions for positioning and navigation can enhance navigation efficiency, which has attracted the attention of researchers. For retrieving corresponding satellite images based on scene text descriptions, Ye et al. \cite{ye2024cross} proposed a text-based localization method, CrossText2Loc, which excels in handling long texts and interpretability. Xia et al. \cite{xia2024uniloc} proposed a Self-Attention Pooling (SAP) module to integrate data from multiple modalities, including natural language, images, and point clouds, to achieve cross-modal place recognition. 
To navigate drones through natural language commands, Chu et al.\cite{chu2024geotext1652} introduced a natural language-guided UAV geolocalization benchmark, GeoText-1652, and proposed a blending spatial matching for region-level spatial relation matching. In addition, Huang et al.\cite{Huang2024vcsr} utilized textual cues through Contextual Region Learning (CRL) and Consistency Semantic Alignment (CSA) mechanisms to guide the model in overcoming challenges related to context understanding and alignment in UAV images.
Unlike existing methods, our approach primarily focuses on addressing the issue of insufficient fine-grained alignment in drone scenarios, which has been overlooked by existing methods.

\subsection{Visual Language Model for Feature Alignment}
Vision-Language Models (VLMs) aim to learn joint representations of images and text. CLIP \cite{radford2021clip} laid the groundwork using large-scale contrastive learning, followed by advancements like UNITER's image-text matching (ITM) \cite{chen2020uniter}, ALBEF's "align-before-fuse" strategy with hard negative mining \cite{li2021albef}, and X-VLM's focus on multi-level concept alignment \cite{zeng2021xvlm}. Architecturally, METER \cite{dou2022meter_swin} assessed various encoders and fusion strategies, while the BLIP series \cite{li2022blip, li2023blip2} employed lightweight modules like Q-Former to integrate understanding and generation tasks efficiently.

\begin{figure*}[t]
  \centering
  \includegraphics[width=\linewidth]{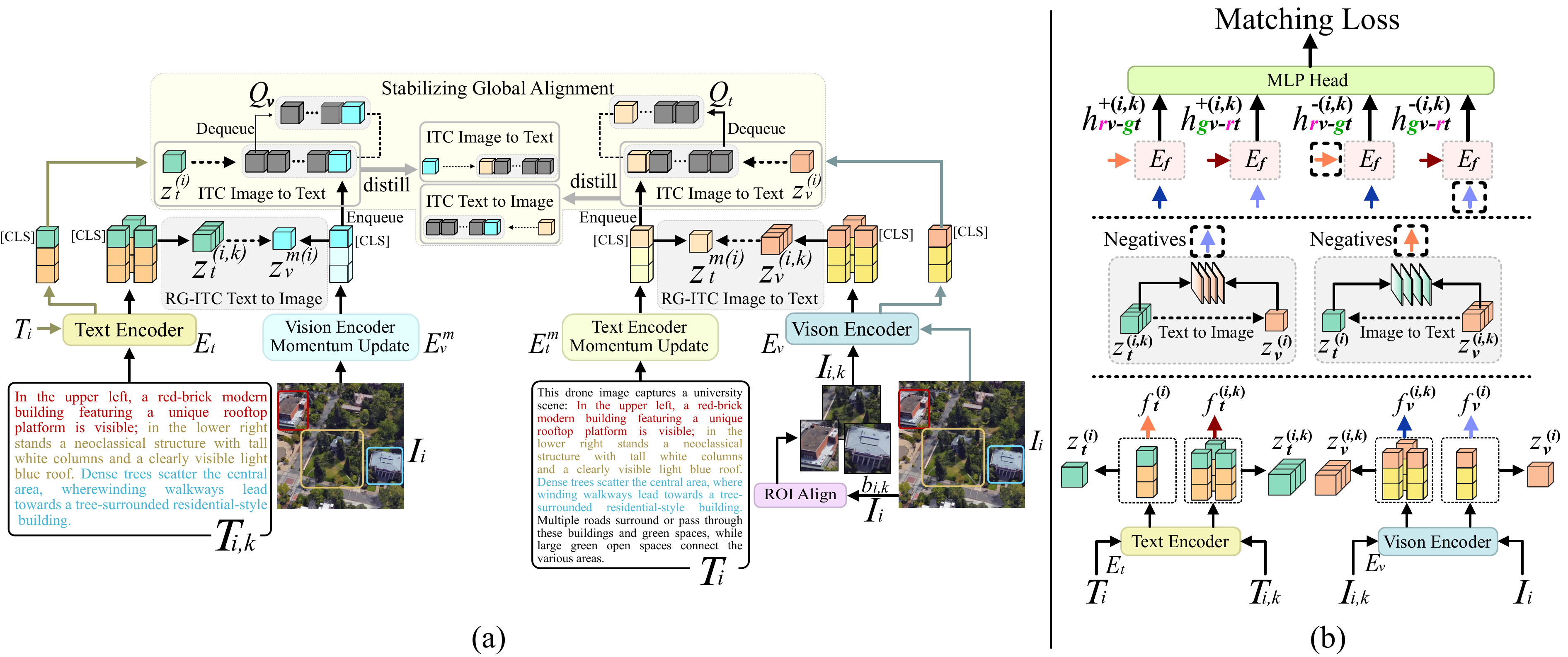}
\caption{Overview of the HCCM architecture. (a) presents the integration of stabilizing global alignment ($\mathcal{L}_{ITC}$) and Region-Global Image Text Contrastive Learning ($\mathcal{L}_{RG-ITC}$), while (b) illustrates Region-Global Image Text Matching Learning ($\mathcal{L}_{RG-ITM}$) module with hard negative mining.}
  \label{fig:overall_method}
\end{figure*}

Standard global image-text alignment fails to capture the hierarchical part-whole concepts inherent in visual and linguistic data. To overcome this, researchers have shifted towards fine-grained and hierarchical approaches. Early methods encoded semantic hierarchies in embedding spaces using partial order constraints or lexical entailment \cite{vendrov2015order, nguyen2017hierarchical, vulic2017specialising}, while others utilized visual structures, aligning text with segmented image regions \cite{arbelaez2010contour, zhang2020self} or focusing on object-level contrastive learning \cite{xie2021unsupervised}.
Recent developments have exploited hyperbolic geometric spaces for hierarchical representation. HyCoCLIP \cite{pal2024compositional}, for instance, models image-text relationships within this space, employing cross-modal contrastive learning between parts and wholes and using the entailment cone loss \cite{ganea2018entailment} to enforce hierarchical constraints both within and across modalities. However, its effectiveness depends on clear part-whole structures or explicit semantic relationships in the data.

\section{Methodology}
\label{sec:methodology}
This section details the proposed Hierarchical Cross-granularity Contrastive and Matching Learning (HCCM) framework. We first outline the base vision-language encoding process (\ref{sec:vl_encoding}).
Subsequently, we introduce the Momentum Contrast and Momentum Distillation (MCD) mechanism employed for stabilizing global alignment (\ref{sec:global_itc_impl}).
Then, we elaborate on the two core components: Region-to-Global Image-Text Contrastive Learning (RG-ITC) for hierarchical semantics learning (\ref{sec:rg_itc}) and Region-Global Image-Text Matching Learning (RG-ITM) for compositional semantics understanding (\ref{sec:rg_itm}). Finally, the overall training objective combining these elements is presented (\ref{sec:overall_loss}). Figure \ref{fig:overall_compare} illustrates the evolution from standard alignment frameworks to the proposed HCCM approach.

\subsection{Vision-Language Encoding}
\label{sec:vl_encoding}
The model processes data in batches of $N$ samples, where each input consists of a global image-text pair $(I_i, T_i)$. Each image $I_i$ is associated with region patches $I_{i,k}$, defined by bounding boxes $b_{i,k}$ and extracted from $I_i$ using ROI Align, along with corresponding text fragments $T_{i,k}$. We adopt XVLM \cite{zeng2021xvlm} as the fundamental architecture, which comprises an image encoder $E_v$, a text encoder $E_t$, and a cross-modal fusion encoder $E_f$.

Global inputs $I_i$ and $T_i$ are encoded by $E_v$ and $E_t$ to produce feature sequences $f_v^{(i)}$ and $f_t^{(i)}$, with their [CLS] token embeddings $f_{v, [\text{CLS}]}^{(i)}$ and $f_{t, [\text{CLS}]}^{(i)}$ serving as global aggregated features. Regional patches $I_{i,k}$ and text fragments $T_{i,k}$ are similarly encoded to yield regional [CLS] embeddings $f_{v, [\text{CLS}]}^{(i,k)}$ and $f_{t, [\text{CLS}]}^{(i,k)}$.

For contrastive learning, all [CLS] token embeddings are mapped through modality-specific projection layers (online: $\phi_v, \phi_t$; momentum: $\phi_v^m, \phi_t^m$) and L2-normalized to generate similarity computation embeddings including global online $z_v^{(i)}, z_t^{(i)}$, global momentum $z_v^{m(i)}, z_t^{m(i)}$, and regional online $z_v^{(i,k)}, z_t^{(i,k)}$.

\subsection{Region-to-Global Image-Text Contrastive Learning}
\label{sec:rg_itc}

Standard Image-Text Contrastive (ITC) learning aims to globally align the semantic representations of image-text pairs, but it overlooks fine-grained cross-modal semantic information. We introduce Region-to-Global Image-Text Contrastive Learning (RG-ITC) to explicitly model part-to-whole cross-modal semantic hierarchical relationships, as illustrated in Figure \ref{fig:overall_method}(a).

Achieving this hierarchical modeling involves contrasting regional online embeddings against global momentum embeddings within a data batch. Specifically, for a batch of $N$ samples, regional pairs $(I_{i,k}, T_{i,k})$ (where $k$ indexes regions in sample $i$) are processed via online encoders ($E_v, E_t$), projection layers ($\phi_v, \phi_t$), and L2 normalization to yield regional online embeddings $z_{v}^{(i,k)}$ and $z_{t}^{(i,k)}$. The contrastive learning objective is then applied: for a regional visual embedding $z_{v}^{(i,k)}$, its positive counterpart is the global textual momentum embedding $z_{t}^{m(i)}$ from the same sample $i$, while the negative counterparts are the global textual momentum embeddings $z_{t}^{m(j)}$ from all other samples $j \neq i$. This process is symmetric for regional text embeddings $z_{t}^{(i,k)}$, which are contrasted against the global visual momentum embedding $z_{v}^{m(i)}$ (positive) and all $z_{v}^{m(j)}$ where $j \neq i$ (negatives).

The RG-ITC loss $\mathcal{L}_{RG-ITC}$ over all valid region pairs $(i,k) \in \mathcal{R}_N$ in the batch is:
\begin{align}
\mathcal{L}_{RG-ITC} =
&-\frac{1}{2|\mathcal{R}_N|} \sum_{(i,k) \in \mathcal{R}_N}
\Biggl[
    \log \frac{\exp(s(z_{v}^{(i,k)}, z_{t}^{m(i)}) / \tau)}{
          \sum\limits_{j=1}^{N}
          \exp(s(z_{v}^{(i,k)}, z_{t}^{m(j)}) / \tau)} \notag \\
&+ \log \frac{\exp(s(z_{t}^{(i,k)}, z_{v}^{m(i)}) / \tau)}{
          \sum\limits_{j=1}^{N}
          \exp(s(z_{t}^{(i,k)}, z_{v}^{m(j)}) / \tau)}
\Biggr], \label{eq:rgitc_final_method_revised}
\end{align}
where $s(\cdot, \cdot)$ denotes cosine similarity, $\tau$ is the temperature, and $N$ is batch size. Minimizing $\mathcal{L}_{RG-ITC}$ fosters learning of local-to-global cross-modal associations.

\subsection{Region-Global Image-Text Matching Learning}
\label{sec:rg_itm}
Image-Text Matching (ITM) is a core task for fine-grained vision-language understanding, typically assessing if a global image-text pair $(I_i, T_i)$ matches. Standard ITM often employs a fusion encoder $E_f$ to combine global features $f_v^{(i)}$ and $f_t^{(i)}$, feeding the fused representation (e.g., from the [CLS] token) into a classification head $H_{match}$ to predict the match probability $p_{match}$. Training minimizes a binary cross-entropy (BCE) loss $\mathcal{L}_{ITM}$ over positive and negative pairs:
\begin{align}
\mathcal{L}_{ITM} &= -\frac{1}{|\mathcal{P}|+|\mathcal{N}|}
\sum_{(I, T) \in \mathcal{P} \cup \mathcal{N}}
\Bigl[ y \log p_{match}(I, T)[1] \notag \\
&\quad + (1-y) \log p_{match}(I, T)[0] \Bigr],
\label{eq:itm_loss_final_prelim_unified_reintro}
\end{align}
where $\mathcal{P}$ and $\mathcal{N}$ are sets of positive and negative pairs, and $y \in \{0,1\}$ is the ground-truth label.

However, standard ITM primarily focuses on global alignment. To better capture local-global consistency and composite semantics (Figure \ref{fig:overall_method}(b)), we introduce Region-Global ITM (RG-ITM). RG-ITM evaluates the alignment between uni-modal local features (regions/fragments) and the cross-modal global representation. Using the shared fusion encoder $E_f$ with global features $f_v^{(i)}, f_t^{(i)}$ and regional features $f_v^{(i,k)}, f_t^{(i,k)}$, we construct positive fused representations by pairing regional features with their corresponding cross-modal global features:
\begin{align}
h_{rv-gt}^{+(i,k)} &= E_f(f_v^{(i,k)}, f_t^{(i)})_{[\text{CLS}]}, \label{eq:pos_h_rvgt_rgitm_revised}\\
h_{gv-rt}^{+(i,k)} &= E_f(f_v^{(i)}, f_t^{(i,k)})_{[\text{CLS}]}. \label{eq:pos_h_gvrt_rgitm_revised}
\end{align}
The set of these positive examples is $\mathcal{H}_{\text{pos}} = \bigl\{ h_{rv-gt}^{+(i,k)},\; h_{gv-rt}^{+(i,k)} \,\bigm|\, (i,k) \in \mathcal{R}_N \bigr\}$, where $\mathcal{R}_N$ covers all valid region pairs $(i,k)$ in the batch.

To improve discrimination, we employ hard negative mining based on online embedding similarity (Figure \ref{fig:overall_method}(b)). For each region $(i,k)$ with online embeddings $z_v^{(i,k)}, z_t^{(i,k)}$, we sample hard negative global counterparts from other samples $j', l' (\neq i)$ with probabilities proportional to their embedding similarity:
\begin{align}
p(j' | i, k) &\propto \exp\Bigl(s(z_v^{(i,k)}, z_t^{(j')}) / \tau\Bigr) \quad \text{for } j' \in \{1..N\}, j' \neq i, \label{eq:neg_sample_prob_t_revised} \\
p(l' | i, k) &\propto \exp\Bigl(s(z_t^{(i,k)}, z_v^{(l')}) / \tau\Bigr) \quad \text{for } l' \in \{1..N\}, l' \neq i. \label{eq:neg_sample_prob_v_revised}
\end{align}
This yields hard negative indices $j^-_{i,k}$ and $l^-_{i,k}$ for each region $(i,k)$.
Negative fused representations are then generated using these sampled indices:
\begin{align}
h_{rv-gt}^{-(i,k)} &= E_f(f_v^{(i,k)}, f_t^{(j^-_{i,k})})_{[\text{CLS}]}, \label{eq:neg_h_rvgt_rgitm_revised}\\[1ex]
h_{gv-rt}^{-(i,k)} &= E_f(f_v^{(l^-_{i,k})}, f_t^{(i,k)})_{[\text{CLS}]}. \label{eq:neg_h_gvrt_rgitm_revised}
\end{align}
The set of negative examples is $\mathcal{H}_{\text{neg}} = \bigl\{ h_{rv-gt}^{-(i,k)},\; h_{gv-rt}^{-(i,k)} \,\bigm|\, (i,k) \in \mathcal{R}_N \bigr\}$.

Finally, we combine positive and negative examples $\mathcal{H}_{\text{RG}} = \mathcal{H}_{\text{pos}} \cup \mathcal{H}_{\text{neg}}$. Each representation $h \in \mathcal{H}_{\text{RG}}$ is processed by the shared matching head $H_{\text{match}}$ to get a probability $p_{\text{match}}^{(h)}$:
\begin{equation}
p_{\text{match}}^{(h)} = \text{softmax}\bigl(H_{\text{match}}(h)\bigr) \in \mathbb{R}^2.
\label{eq:match_prob_rgitm_revised}
\end{equation}
The RG-ITM loss $\mathcal{L}_{\text{RG-ITM}}$ minimizes the BCE over all examples in $\mathcal{H}_{\text{RG}}$:
\begin{align}
\mathcal{L}_{\text{RG-ITM}} =
-\frac{1}{|\mathcal{H}_{\text{RG}}|} \sum_{h \in \mathcal{H}_{\text{RG}}}
\Bigl[\, & y^{(h)} \log p_{\text{match}}^{(h)}[1] \nonumber\\
&+ (1-y^{(h)}) \log p_{\text{match}}^{(h)}[0] \Bigr],
\label{eq:rg_itm_loss_final_method_revised}
\end{align}
where $y^{(h)}$ is 1 for positive examples and 0 for negative ones.

\subsection{Stabilizing Global Alignment with Momentum Contrast and Distillation}
\label{sec:global_itc_impl}
However, employing only the hierarchical strategies encounters certain limitations within drone scenarios. Due to the large scale of drone bird's-eye views, textual descriptions ($T_i$ and $T_{i,k}$) are often incomplete or ambiguous. When faced with this situation, the aforementioned methods, during the process of cross-modal local-to-global information alignment, can amplify the potential negative effects stemming from local descriptive inaccuracies or omissions. This, in turn, impacts the model's crucial global alignment.

To address this issue, we introduce a dual technique applied to the global contrastive learning between $I_i$ and $T_i$: Momentum Contrast and Momentum Distillation (MCD) (Figure \ref{fig:overall_method}(a)). These utilize online ($\theta$) and momentum ($\theta^m$, updated via Exponential Moving Average) encoders.

\textbf{Momentum Contrast.} Inspired by MoCo \cite{he2020momentum}, we employ momentum queues ($Q_v, Q_t$) storing historical global momentum features ($z_v^m, z_t^m$). The resulting large, stable negative sets ($Z_v^m = \{z_v^{m(j)}\}_{j=1}^N \cup Q_v$, $Z_t^m = \{z_t^{m(j)}\}_{j=1}^N \cup Q_t$) enhance the model's discriminative ability in noisy data, forcing it to focus on true global differences rather than potentially amplified local noise signals.

\textbf{Momentum Distillation.} Inspired by ALBEF \cite{li2021align}, we generate soft targets ($q^{(i)}$) for global ITC by blending momentum model predictions with ground-truth labels ($y^{(i)}$) using coefficient $\alpha$ and temperature parameter $\tau$:
\begin{align}
q_{i2t}^{(i)} &= \alpha \cdot \text{softmax}(z_v^{m(i) \top} Z_t^m / \tau) + (1-\alpha) \cdot y_{i2t}^{(i)} \notag \\
q_{t2i}^{(i)} &= \alpha \cdot \text{softmax}(z_t^{m(i) \top} Z_v^m / \tau) + (1-\alpha) \cdot y_{t2i}^{(i)} \label{eq:distill_targets_method_unified}
\end{align}
Momentum distillation, through temporally smoothed model parameters, constructs a trend-based supervisory signal for the model's learning process, guiding the online encoder to resist noise interference.

The stabilized global ITC loss using MCD, denoted as $\mathcal{L}_{ITC_{MCD}}$, employs cross-entropy ($H(\cdot, \cdot)$) between online global predictions and these soft targets:
\begin{equation}
\begin{aligned}
\mathcal{L}_{ITC_{MCD}} = &-\frac{1}{2N} \sum_{i=1}^N \Big[
H\left(q_{i2t}^{(i)}, \text{softmax}(z_v^{(i) \top} Z_t^m / \tau)\right) \\
&+ H\left(q_{t2i}^{(i)}, \text{softmax}(z_t^{(i) \top} Z_v^m / \tau)\right) \Big]
\end{aligned}
\label{eq:itc_mcd_loss_final_method_unified} 
\end{equation}

This approach, by stabilizing the foundational global alignment, enhances model performance and robustness when handling local-to-global information in challenging drone environments.

\subsection{Overall Training Objective}
\label{sec:overall_loss}

HCCM's overall training objective $\mathcal{L}_{total}$ combines multiple losses to jointly optimize hierarchical, part-to-whole cross-modal alignment and learn complex compositional semantics, particularly those involving relational structures in visual scenes.
Consistent with standard practices \cite{zeng2021xvlm, chu2024geotext1652}, we incorporate a bounding box regression loss, $\mathcal{L}_{Box}$, to further refine the model's region localization capabilities. This loss penalizes the difference between the predicted bounding box $\hat{b}_k$ (regressed from fused features via head $H_{box}$) and the ground-truth box $b_{i,k}$ corresponding to text fragment $T_{i,k}$, using a combination of L1 distance and GIoU loss \cite{rezatofighi2019generalized}:
\begin{align}
\mathcal{L}_{Box} &= \sum_{k} \big[ \lambda_{L1} ||\hat{b}_k - b_{i,k}||_1 + \lambda_{GIoU} (1 - \text{GIoU}(\hat{b}_k, b_{i,k})) \big],
\label{eq:box_loss_revised_prelim_unified}
\end{align}
where $\lambda_{L1}$ and $\lambda_{GIoU}$ are respective weights.

The complete training objective $\mathcal{L}_{total}$ for the HCCM framework is the weighted sum of all constituent losses:
\begin{align}
\mathcal{L}_{total} =
&w_{itc} \mathcal{L}_{ITC_{MCD}} + w_{itm} \mathcal{L}_{ITM} \notag \\ 
&+ w_{rg-itc} \mathcal{L}_{RG-ITC} + w_{rg-itm} \mathcal{L}_{RG-ITM} \notag \\
&+ w_{box} \mathcal{L}_{Box}.
\label{eq:total_loss_final_method_unified}
\end{align}
This objective combines the stabilized global contrastive loss with MCD ($\mathcal{L}_{ITC_{MCD}}$, Eq. \ref{eq:itc_mcd_loss_final_method_unified}), the standard global matching loss ($\mathcal{L}_{ITM}$, Eq. \ref{eq:itm_loss_final_prelim_unified_reintro}), the proposed region-global contrastive ($\mathcal{L}_{RG-ITC}$, Eq. \ref{eq:rgitc_final_method_revised}) and matching ($\mathcal{L}_{RG-ITM}$, Eq. \ref{eq:rg_itm_loss_final_method_revised}) objectives, along with the bounding box regression term ($\mathcal{L}_{Box}$, Eq. \ref{eq:box_loss_revised_prelim_unified}). Each component's contribution is balanced by its weight $w_{(\cdot)}$. Minimizing $\mathcal{L}_{total}$ guides the HCCM model to effectively learn hierarchical, multi-granularity vision-language representations by integrating both global context and local details.

\section{Experiments}
\label{sec:experiment}
This section evaluates the proposed HCCM method on the Natural Language-Guided Drone (NLGD) tasks of UAV text navigation (text-to-image retrieval) and UAV view target localization (image-to-text retrieval). We first outline the experimental setup, detailing the datasets, metrics (\ref{subsec:datasets_metrics}), and implementation (\ref{subsec:implementation}). We then present comparative results against state-of-the-art methods (\ref{subsec:comparison_sota}), followed by ablation studies (\ref{subsec:ablation}) and an assessment of zero-shot generalization (\ref{subsec:zeroshot_eval}). Finally, visualization is provided (\ref{subsec:qualitative}). 

\subsection{Datasets and Evaluation Metrics}
\label{subsec:datasets_metrics}
We conduct training and primary evaluation using the GeoText-1652 dataset \cite{chu2024geotext1652}, strictly adhering to its official data splits and evaluation protocols. For assessing zero-shot generalization, we additionally evaluate performance on the ERA dataset \cite{Huang2024vcsr}.

Across all experiments, performance is measured using Recall@K metrics, specifically Recall@1 (R@1), Recall@5 (R@5), and Recall@10 (R@10). In the zero-shot generalization evaluation, we additionally use mean Recall (mR).

\subsection{Implementation Details}
\label{subsec:implementation}
For fair comparison, we follow the setup from the GeoText-1652 \cite{chu2024geotext1652}, which uses a standard XVLM \cite{zeng2021xvlm} model pre-trained on 16 million image-text pairs as the backbone. We fine-tune our model on GeoText-1652 dataset and employ the AdamW \cite{loshchilov2017decoupled} optimizer (initial learning rate $3 \times 10^{-5}$, weight decay $0.01$) for 6 epochs with batch size 24. 
Key hyperparameters for our HCCM method are as follows: momentum $\beta=0.995$, queue size $Q=57,600$, distillation $\alpha=0.4$, and temperature $\tau=0.07$. Loss weights, determined via preliminary search, are $w_{itc}=0.25$, $w_{itm}=1$, $w_{rg-itc}=0.25$, $w_{rg-itm}=0.5$, and $w_{box}=0.1$.

\subsection{Comparison with State-of-the-art Methods}
\label{subsec:comparison_sota}

In this experiment, we compare our HCCM with existing competitive methods on the GeoText-1652 dataset under both zero-shot and fine-tuned settings. We report the results of R@1, R@5, and R@10 across all methods, alongside the model parameters and the pretrained image size used. Notably, we reproduce HyCoCLIP \cite{pal2024compositional} on the NLGD task using publicly available code.

\begin{table}[t]
  \centering
  \small
  \caption{Comparative performance evaluation of cross-modal retrieval methods on the GeoText-1652 benchmark. Results are presented using Recall@K (R@K) for both Image Query (Drone-view Geolocalization) and Text Query (Drone Navigation) tasks, under zero-shot and fine-tuned settings. $\dagger$ denotes results are reproduced by the provided source code. The best performances are in \textbf{bold}.} 
  \setlength{\tabcolsep}{1.2pt} 
  \label{tab:comparison_results_adjusted_v2} 
  \begin{tabular*}{\linewidth}{@{\extracolsep{\fill}} l c c r r r r r r @{}} 
    \toprule
    \multirow{2}{*}{Method} & \multirow{2}{*}{Params} & \multirow{2}{*}{Pretrained}
    & \multicolumn{3}{c}{Image Query(\%)} & \multicolumn{3}{c}{Text Query(\%)} \\
    \cmidrule(lr){4-9}
    & &Images & R@1 & R@5 & R@10 & R@1 & R@5 & R@10 \\
    \midrule
    \multicolumn{9}{c}{\textit{Zero-Shot Evaluation on GeoText-1652}} \\
    UNITER \cite{chen2020uniter}       & 300M & 4M  & 2.5  & 7.4  & 11.8 & 0.9  & 2.7  & 4.2  \\
    METER-Swin \cite{dou2022meter_swin}   & 380M & 4M  & 2.7  & 8.0  & 12.2 & 1.3  & 3.9  & 5.8  \\
    ALBEF \cite{li2021albef}        & 210M & 4M  & 2.9  & 8.1  & 12.4 & 1.8  & 4.8  & 7.1  \\
    ALBEF \cite{li2021albef}        & 210M & 14M & 3.0  & 9.1  & 14.2 & 1.1  & 3.5  & 5.3  \\
    XVLM \cite{zeng2021xvlm}         & 216M & 4M  & 4.9  & 14.2 & 21.1 & 4.3  & 9.1  & 13.2 \\
    XVLM \cite{zeng2021xvlm}         & 216M & 16M & 5.0  & 14.4 & 21.4 & 4.5  & 9.9  & 13.4 \\
    \midrule
    \multicolumn{9}{c}{\textit{Fine-Tuned Evaluation on GeoText-1652}} \\
    HyCoCLIP \cite{pal2024compositional}$\dagger$      & 216M & 16M & 15.3 & 33.6 & 43.2 & 8.7  & 15.8 & 20.0 \\
    UNITER \cite{chen2020uniter}       & 300M & 4M  & 21.4 & 43.4 & 59.5 & 10.6 & 20.4 & 26.1 \\
    METER-Swin \cite{dou2022meter_swin}   & 380M & 4M  & 22.7 & 46.3 & 60.7 & 11.3 & 21.5 & 27.3 \\
    ALBEF \cite{li2021albef}        & 210M & 4M  & 22.9 & 49.5 & 62.3 & 12.3 & 22.8 & 28.6 \\
    ALBEF \cite{li2021albef}        & 210M & 14M & 23.2 & 49.7 & 62.4 & 12.5 & 22.8 & 28.5 \\
    XVLM \cite{zeng2021xvlm}         & 216M & 4M  & 23.6 & 50.0 & 63.2 & 13.1 & 23.5 & 29.2 \\
    XVLM \cite{zeng2021xvlm}         & 216M & 16M & 25.0 & 52.3 & 65.1 & 13.2 & 23.7 & 29.6 \\
    GeoText-1652 \cite{chu2024geotext1652}      & 217M & 16M & 26.3 & 53.7 & 66.9 & 13.6 & 24.6 & 31.2 \\
    HCCM       & 216M & 16M & \textbf{28.8} & \textbf{57.3} & \textbf{69.9} & \textbf{14.7} & \textbf{26.0} & \textbf{32.5} \\
    \bottomrule
  \end{tabular*}
\end{table}

From the results shown in Table \ref{tab:comparison_results_adjusted_v2}, we can observe that: 
1) In both Image Query and Text Query settings, our method achieves the best performance, significantly outperforming other methods across all metrics. Specifically, we reach 28.8\% R@1 in the Image Query setting and 14.7\% R@1 in the Text Query setting.
2) Compared to the state-of-the-art method Geotext-1652 \cite{chu2024geotext1652} of the NLGD task, our method models the hierarchical relationships across modalities and performs fine-grained features alignment, leading to superior performance.
3) HyCoCLIP \cite{pal2024compositional} employs compositional entailment learning to model the part-whole hierarchical relationships, which is greatly limited in drone scenarios. In contrast, our method utilizes the proposed RG-ITC and RG-ITM learning strategies, which are better at extracting the complex, intertwined spatial semantic information present in drone scenes.
4) Compared to Text Query setting, all methods perform better in Image Query setting, indicating that the text retrieval task are more challenging.


\begin{table}[t]
  \centering
  \small
  \caption{Ablation study on the GeoText-1652 dataset evaluating the contribution of individual components of our proposed HCCM method. 
  \textit{MC} and \textit{MD} denote the proposed momentum contrast and momentum distillation, while \textit{RG-ITC} and \textit{RG-ITM} represent our region-global image text contrastive learning and region-global image text matching learning.
   } 
  \label{tab:pro_compact_layout_adjusted} 
  \setlength{\tabcolsep}{2pt} 
  \begin{tabular}{cccc rrr rrr} 
    \toprule
    \multicolumn{4}{c}{Components} & 
    \multicolumn{3}{c}{Image Query(\%)} &
    \multicolumn{3}{c}{Text Query(\%)} \\
    \cmidrule(lr){1-4} \cmidrule(lr){5-7} \cmidrule(lr){8-10} 
    \textit{MC} & \textit{MD} & \textit{RG-ITC} & \textit{RG-ITM} & R@1 & R@5 & R@10 & R@1 & R@5 & R@10 \\
    \midrule
     &  &  &  & 25.51 & 52.70 & 65.54 & 12.84 & 23.07 & 29.27 \\ 
    $\checkmark$ &  &  &  & 26.48 & 54.10 & 66.78 & 13.75 & 24.91 & 31.49 \\ 
    $\checkmark$ & $\checkmark$ &  &  & 26.86 & 54.96 & 67.83 & 14.11 & 25.13 & 31.77 \\ 
     &  & $\checkmark$ &  & 27.01 & 54.63 & 67.05 & 13.63 & 24.24 & 30.50 \\ 
     &  & $\checkmark$ & $\checkmark$ & 27.04 & 54.91 & 67.52 & 14.04 & 24.69 & 30.97 \\ 
    $\checkmark$ & $\checkmark$ & $\checkmark$ &  & 27.32 & 55.78 & 68.53 & 14.15 & 25.21 & 31.81 \\ 
    $\checkmark$ & $\checkmark$ &  & $\checkmark$ & 26.89 & 54.86 & 67.72 & 14.41 & 25.77 & 32.28 \\ 
    $\checkmark$ & $\checkmark$ & $\checkmark$ & $\checkmark$ & \textbf{28.82} & \textbf{57.30} & \textbf{69.93} & \textbf{14.73} & \textbf{25.98} & \textbf{32.49} \\ 
    \bottomrule
  \end{tabular}
\end{table}

\begin{table}[t]
\centering
\small
\setlength{\tabcolsep}{2pt}
\caption{Ablation study of internal components in RG-ITC and RG-ITM to explore the impact of removing cross-modal directional losses on the performance of models.}
\label{tab:ablation_rg_components}
\begin{tabular}{@{}lcccccc@{}} 
\toprule 
\multicolumn{1}{c}{\multirow{2}{*}{Method}} & \multicolumn{3}{c}{Image Query (\%)} & \multicolumn{3}{c}{Text Query (\%)} \\ 
\cmidrule(lr){2-4} \cmidrule(lr){5-7} 
                         & R@1   & R@5   & R@10  & R@1   & R@5   & R@10  \\
\midrule 
HCCM                    & 28.82 & 57.30 & 69.93 & 14.73 & 25.98 & 32.49 \\
$- \mathcal{L}_{RG-ITC}(I_{i,k}\rightarrow T_i)$ & 27.62 & 56.22 & 68.90 & 14.43 & 25.46 & 32.02 \\
$- \mathcal{L}_{RG-ITC}(T_{i,k} \rightarrow I_i) $ & 26.89 & 54.86 & 67.72 & 14.41 & 25.77 & 32.28 \\
$- \mathcal{L}_{RG-ITM}(I_{i,k}\leftrightarrow T_i)$ & 27.19 & 55.37 & 68.33 & 14.47 & 25.37 & 31.80 \\
$- \mathcal{L}_{RG-ITM}(T_{i,k} \leftrightarrow I_i)$ & 26.86 & 54.96 & 67.83 & 14.11 & 25.13 & 31.77 \\
\bottomrule 
\end{tabular}
\end{table}

\begin{table}[t]
\centering
\small
\caption{Comparison of fine-tuned results and zero-shot results on the ERA dataset. \textit{mR} denotes mean Recall.
}
\label{tab:era_comparison_final}
\setlength{\tabcolsep}{3pt}
\begin{tabular}{@{}l llllll l@{}}
\toprule
\multirow{2}{*}{Method} & \multicolumn{3}{c}{Image Retrieval (\%)} & \multicolumn{3}{c}{Text Retrieval (\%)} & \multirow{2}{*}{mR (\%)} \\
\cmidrule(lr){2-7}
& R@1 & R@5 & R@10 & R@1 & R@5 & R@10 & \\
\midrule
\multicolumn{8}{c}{\textit{Reported Fine-tuned Results on ERA Dataset}} \\

VSE++ \cite{faghri2017vsepp}         & 10.13 & 35.20 & 53.91 & 9.79 & 30.40 & 42.90 & 30.39 \\
PVSE K=2 \cite{song2019polysemous}      & 11.04 & 35.57 & 51.65 & 11.31 & 32.60 & 46.95 & 31.52 \\
PVSE K=1 \cite{song2019polysemous}      & 11.14 & 36.08 & 53.75 & 9.96 & 33.95 & 47.97 & 32.14 \\
CLIP \cite{radford2021learning}          & 12.73 & 37.33 & 51.52 & 11.31 & 31.92 & 43.91 & 31.45 \\
PCME \cite{chun2021probabilistic}          & 13.85 & 42.87 & 60.64 & 14.69 & 35.30 & 49.15 & 36.08 \\
AMFMN-soft \cite{yuan2022exploring}    & 14.18 & \textbf{46.79} & 62.87 & 14.35 & 38.01 & 52.02 & 38.04 \\
AMFMN-sim \cite{yuan2022exploring}     & 13.75 & 43.41 & 59.59 & 14.02 & 34.12 & 51.52 & 36.06 \\
AMFMN-fusion \cite{yuan2022exploring}  & 11.62 & 42.26 & 60.51 & 15.20 & 36.99 & 50.33 & 36.15 \\
GALR \cite{yuan2022remote}          & 14.03 & 45.15 & 64.54 & 12.38 & 36.59 & 50.90 & 37.27 \\
VCSR \cite{Huang2024vcsr}           & 13.69 & 46.31 & \textbf{66.37} & 15.65 & 38.28 & 53.49 & 38.96 \\
\midrule
\multicolumn{8}{c}{\textit{Zero-shot Results on ERA Dataset (Fine-tuned on GeoText-1652)}} \\
HyCoCLIP \cite{pal2024compositional}  & 7.77  & 17.57 & 23.31 & 8.04  & 18.65 & 21.96 & 16.22 \\
XVLM \cite{zeng2021xvlm}    & 14.19 & 36.15 & 50.34 & 14.05 & 39.46 & 53.99 & 34.70 \\
GeoText-1652 \cite{chu2024geotext1652}  & 17.91 & 39.19 & 54.73 & 17.09 & 42.30 & 56.76 & 38.00 \\
HCCM   & \textbf{19.93} & 39.19 & 56.76 & \textbf{18.58} & \textbf{45.20} & \textbf{59.93} & \textbf{39.93} \\
\bottomrule
\end{tabular}
\end{table}

\subsection{Ablation Study}
\label{subsec:ablation}
We perform ablation studies on the GeoText-1652 \cite{chu2024geotext1652} dataset to evaluate the effectiveness of individual components of the proposed HCCM. Refer to \cite{chu2024geotext1652}, the standard XVLM \cite{zeng2021xvlm} is adopted as our baseline. Results are presented in Table \ref{tab:pro_compact_layout_adjusted} and Table \ref{tab:ablation_rg_components}.

As the result shown in Table \ref{tab:pro_compact_layout_adjusted}, we evaluate the contribution of individual components of our proposed HCCM method. When incorporating only the momentum contrast \textit{MC} and the momentum distillation \textit{MD} (row 3), the R@1 of the baseline model can be improved from 25.51\% to 26.86\% (+1.35 points) in the Image Query setting, indicating the usefulness of enhancing global representation stability. 
Similarly, when integrating only the cross-granularity learning components (i.e., \textit{RG-ITC} and \textit{RG-ITM}), the R@1 of the model can be raised to 27.04\% (+1.53 points), confirming the benefit of capturing fine-grained information. 
By combining all components, our proposed HCCM can achieve the highest performance with an Image Query R@1 of 28.82\% and a Text Query R@1 of 14.73\%, which surpasses the momentum-only configuration (row 3) by 1.96 points and the cross-granularity-only setup (row 5) by 1.78 points in Image Query R@1. The above results highlight a significant synergy: robust global alignment provides a stable foundation, while cross-granularity learning contributes essential fine-grained details, leading to optimal performance when combined.


Further analysis in Table \ref{tab:ablation_rg_components} investigates the impact of directional losses within RG-ITC and RG-ITM. Compared to the HCCM model (row 1), removing any single directional loss component results in performance degradation. Notably, excluding the text-region to global image association in RG-ITC (i.e., $- \mathcal{L}_{RG-ITC}(T_{i,k}\rightarrow I_i) $ in row 3), the Image Query R@1 of the model is reduced by 1.93 points to 26.89\%. Similarly, removing the text-region to global image matching in RG-ITM (i.e., $- \mathcal{L}_{RG-ITM}(T_{i,k} \leftrightarrow I_i)$ in row 5), a drop of 1.96 points to 26.86\% is caused. This suggests that learning text-to-image associations is particularly crucial, and confirms that all proposed bidirectional cross-modal learning components contribute positively to the overall performance.

\begin{figure*}[t]
  \centering
  \includegraphics[width=0.87\linewidth]{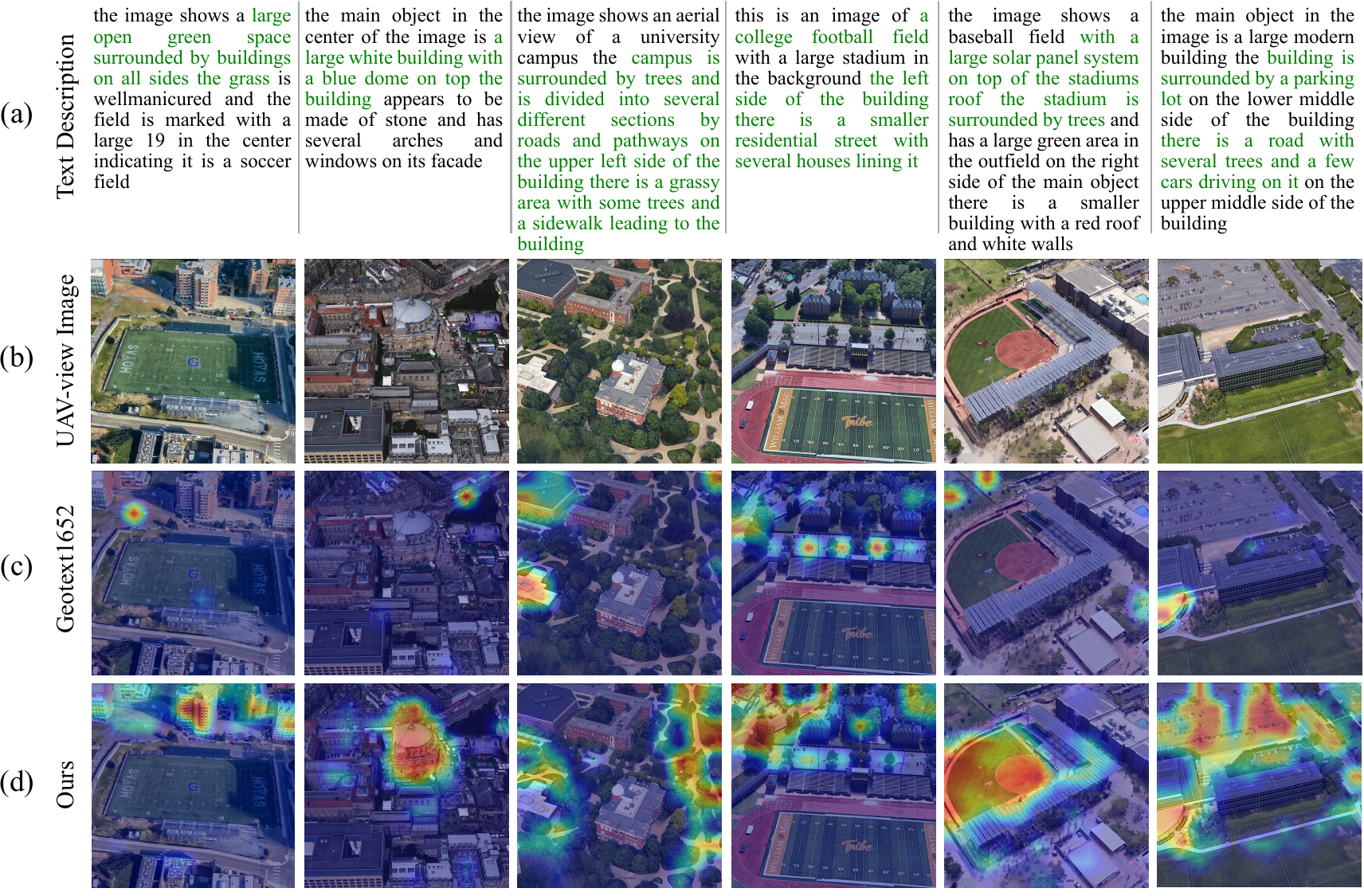}
   \caption{Visualization of activation maps. (a) Text descriptions with \textbf{key descriptions} highlighted in green. (b) Corresponding UAV-view images. (c) Activation maps from the SOTA method (GeoText-1652 \cite{chu2024geotext1652}). (d) Activation maps from our proposed HCCM method.}
  \label{fig:visualization_discussion}
\end{figure*}

\subsection{Zero-shot Generalization Evaluation}
\label{subsec:zeroshot_eval}
We assessed generalization via zero-shot cross-modal retrieval on the unseen ERA dataset \cite{Huang2024vcsr}, using models fine-tuned on GeoText-1652 \cite{chu2024geotext1652}. Table \ref{tab:era_comparison_final} compares HCCM with other competitive methods fine-tuned on GeoText-1652 \cite{chu2024geotext1652, pal2024compositional, zeng2021xvlm} and benchmark models fine-tuned directly on ERA \cite{faghri2017vsepp, song2019polysemous, radford2021learning, chun2021probabilistic, yuan2022exploring, yuan2022remote, Huang2024vcsr}.

In zero-shot evaluation (Table \ref{tab:era_comparison_final}, bottom), HCCM surpasses all methods fine-tuned only on GeoText-1652, achieving state-of-the-art R@1 (19.93\% in image retrieval, 18.58\% in text retrieval) and mR (39.93\%). This notably exceeds the suboptimal GeoText-1652 \cite{chu2024geotext1652} (+2.02\% image R@1, +1.49\% text R@1, +1.93\% mR). Crucially, the zero-shot performance of HCCM on ERA even exceeds that of models fine-tuned directly on ERA (Table \ref{tab:era_comparison_final}, top). The zero-shot mR (39.93\%) of HCCM surpasses the best fine-tuned mR (38.96\% by VCSR \cite{Huang2024vcsr}). Likewise, the zero-shot R@1 scores of HCCM outperform the best respective fine-tuned R@1 results (14.18\% image \cite{yuan2022exploring}, 15.65\% text \cite{Huang2024vcsr}). 
The above results demonstrate that the hierarchical cross-granularity learning strategy of HCCM leads to exceptional zero-shot generalization capabilities. By generating robust transferable representations, HCCM effectively models compositional semantics to achieve powerful generalization, even surpassing models fine-tuned on the target domain.

\subsection{Visualizing Attention for Semantic Grounding}
\label{subsec:qualitative}

We compare GradCAM \cite{selvaraju2017grad} activation maps (Figure~\ref{fig:visualization_discussion}) for HCCM (row d) and SOTA GeoText-1652 \cite{chu2024geotext1652} (row c) to assess fine-grained and compositional grounding against text descriptions (row a).

The SOTA method struggles to accurately ground fine-grained entities (e.g., "blue dome" col 2, "solar panels" col 5) and compositional relationships (e.g., "building surrounded by" col 1, "parking lot"/"road" layout col 6). Its diffuse attention in complex scenes (cols 3, 4) suggests difficulty parsing intricate arrangements, possibly from over-reliance on global matching. Conversely, HCCM shows significantly improved semantic grounding, localizing fine-grained details ("blue dome" col 2, "solar panels" col 5) and better capturing compositional semantics: relating fields to "surrounding buildings" (col 1), reflecting campus structure via "roads and pathways" (col 3), linking fields to "residential streets" (col 4), and grounding the building/"parking lot"/"road" context (col 6). This improved relational grounding, consistent with RG-ITC and RG-ITM objectives, is likely aided by RG-ITM's consistency evaluation. Notably, in scenes with large uniform areas (e.g., fields in cols 1, 3), neither model strongly activates the primary object. This might occur because distinguishing such scenes relies more on grounding discriminative textual descriptions of contrasting features, boundaries, or relationships than on the homogenous central region.

In summary, HCCM's hierarchical modeling (RG-ITC) and consistency evaluation (RG-ITM) yield more precise semantic grounding than global alignment methods alone. Its enhanced relational interpretation benefits complex NLGD tasks, especially in drone-view scenarios.


\section{Conclusion}
\label{sec:conclusion}
This paper introduces HCCM to enhance fine-grained and compositional understanding in Natural Language-Guided Drone (NLGD) tasks. By integrating Region-Global Contrastive (RG-ITC) and Matching (RG-ITM) learning, HCCM effectively models hierarchical cross-modal semantics and evaluates local-global consistency without requiring strict entity partitioning. Furthermore, a Momentum Contrast and Distillation (MCD) mechanism stabilizes global alignment against ambiguous descriptions common in drone views. Extensive experiments validate HCCM's effectiveness, demonstrating state-of-the-art performance on the GeoText-1652 benchmark and superior zero-shot generalization on the ERA dataset, highlighting its robustness for complex drone vision-language applications.

\begin{acks}
This work is supported by the National Natural Science Foundation of China under Grant No.~62276221,~62376232; the Open Project Program of Fujian Key Laboratory of Big Data Application and Intellectualization for Tea Industry, Wuyi University (No. FKLBDAITI202401).
\end{acks}

\bibliographystyle{ACM-Reference-Format}
\balance
\bibliography{sample-base}

\end{document}